\documentclass{article}
\usepackage{eeg}

\usepackage{times}
\usepackage{soul}
\usepackage{url}
\usepackage[hidelinks]{hyperref}
\usepackage[utf8]{inputenc}
\usepackage[small]{caption}
\usepackage{graphicx}
\usepackage{amsmath}
\usepackage{amsthm}
\usepackage{booktabs}
\usepackage{algorithm}
\usepackage{algorithmic}
\usepackage[switch]{lineno}

\usepackage{xspace}
\usepackage{color}
\usepackage{enumitem}
\usepackage{subcaption}
\usepackage{multirow}
\usepackage{multicol}
\usepackage{diagbox}
\usepackage{makecell}

\newcommand{\our}{\textsc{EEG2Text}\xspace}

\title{\our: Open Vocabulary EEG-to-Text Decoding with EEG Pre-Training and Multi-View Transformer}

\author{${^1}$Hanwen Liu, ${^1}$Daniel Hajialigol, ${^2}$Benny Antony, ${^3}$Aiguo Han, ${^1}$Xuan Wang \\
  ${^1}$ Department of Computer Science, Virginia Tech, VA, USA \\
  ${^2}$ Department of Electrical and Computer Engineering, Virginia Tech, VA, USA \\
  ${^3}$ Department of Biomedical Engineering and Mechanics, Virginia Tech, VA, USA \\  
  \texttt{\{liuhwen,danielhajialigol,bennyantony,aiguohan,xuanw\}@vt.edu} \\
  }

\begin{document}

\maketitle

\begin{abstract}
Deciphering the intricacies of the human brain has captivated curiosity for centuries. Recent strides in Brain-Computer Interface (BCI) technology, particularly using motor imagery, have restored motor functions such as reaching, grasping, and walking in paralyzed individuals. However, unraveling natural language from brain signals remains a formidable challenge. Electroencephalography (EEG) is a non-invasive technique used to record electrical activity in the brain by placing electrodes on the scalp. Previous studies of EEG-to-text decoding have achieved high accuracy on small closed vocabularies, but still fall short of high accuracy when dealing with large open vocabularies.  
We propose a novel method, \our, to improve the accuracy of open vocabulary EEG-to-text decoding. 
Specifically, \our leverages EEG pre-training to enhance the learning of semantics from EEG signals and proposes a multi-view transformer to model the EEG signal processing by different spatial regions of the brain. Experiments show that \our has superior performance, outperforming the state-of-the-art baseline methods by a large margin of up to 5\% in absolute BLEU and ROUGE scores. \our shows great potential for a high-performance open-vocabulary brain-to-text system to facilitate communication. 
\end{abstract}

\section{Introduction}

Recent advances in brain-computer interface (BCI) technology have demonstrated exciting progress in restoring the capabilities of patients with paralysis, particularly restoring such motor functions as reaching \cite{hochberg2012reach}, grasping \cite{aflalo2015decoding,bouton2016restoring}, and walking \cite{lorach2023walking}. The heart of BCI is its ability to accurately decode complex brain signals. Despite the advances in decoding brain signals related to motion, decoding brain signals related to speech remains a formidable challenge. Previous research translating speech-related brain signals to text (brain-to-text) primarily relies on electrocorticography (ECoG), an invasive electrophysiological monitoring method that uses electrodes placed directly on the exposed brain surface to record activity from the cerebral cortex. ECoG offers higher temporal and spatial resolution than traditional noninvasive scalp electroencephalography (EEG), with a significantly better signal-to-noise ratio. However, the invasive nature of ECoG is undesirable for BCI applications, and it is highly desirable to develop brain-to-text decoding methods using noninvasive EEG signals, although EEG signals are significantly more challenging to work with than ECoG.  

Previous studies of EEG-to-text decoding \cite{herff2015brain,sun2019towards,anumanchipalli2019speech,makin2020machine,panachakel2021decoding,moses2021neuroprosthesis,nieto2022thinking} have achieved high accuracy on small closed vocabularies, but still fall short of high accuracy when dealing with large open vocabularies. These approaches primarily target high accuracy ($>90\%$) but are often confined to small closed vocabularies and struggle to decode semantically similar words beyond training sets. Recent studies broaden the scope from closed to open-vocabulary EEG-to-text decoding \cite{DBLP:journals/corr/abs-2112-02690,willett2023high,tang2023semantic,duan2023dewave}, drastically expanding the vocabulary size by over 100-fold, from several hundred to tens of thousands of words. Notably, two of these studies \cite{DBLP:journals/corr/abs-2112-02690,duan2023dewave} leverage a pre-trained large language model BART \cite{bart}, and represent the state-of-the-art for open vocabulary brain-to-text decoding. However, these studies are in their nascent stages and are challenged by their limited accuracy.

\begin{figure*}[t]
\centering
\includegraphics[width=\linewidth]{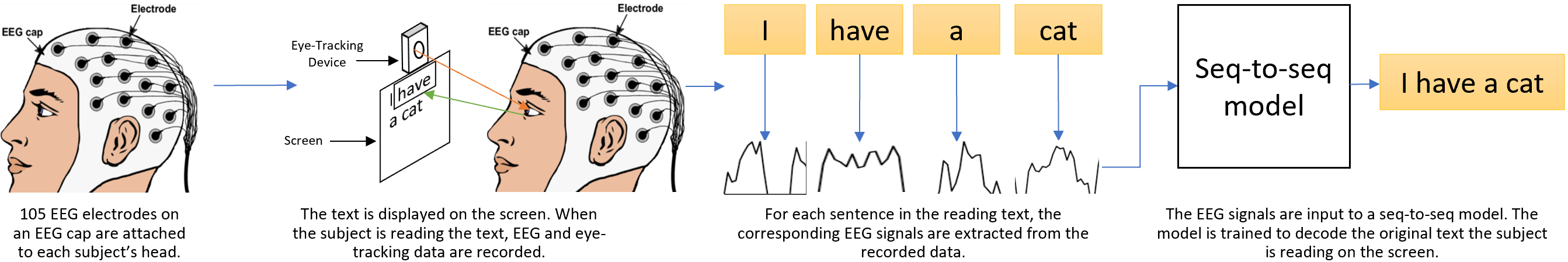}
\caption {The overall framework of open-vocabulary EEG-to-text translation. The first sub-figure comes from \protect\cite{nagel2018modelling}.}
\label{fig:eeg_task}
\end{figure*}

To improve the accuracy of EEG-to-text decoding with open vocabularies, we propose a novel EEG-to-text decoding method based on transformers. First, we introduce a Convolutional Neural Network (CNN) module before the base transformer model to enhance the model's ability to handle long EEG signals. Second, we conduct pre-training of the transformer model by reconstructing randomly masked EEG signals from the input data. This pre-training step helps our transformer model better learn the semantics of EEG signals. Last, we propose a multi-view transformer architecture, where each single-view transformer is the pre-trained model from the previous step, to model the EEG signal processing by different spatial regions of the brain.
Experiments show that \our has superior performance, outperforming the state-of-the-art baseline methods by a large margin of up to 5\% in absolute BLEU and ROUGE scores. \our shows great potential for a high-performance open-vocabulary brain-to-text system to facilitate communication. We will open-source our code and dataset to facilitate future studies of EEG-to-text translation.

\section{Task Definition}

Our task involves decoding corresponding natural language text from raw EEG signals (Figure \ref{fig:eeg_task}). The data acquisition process involves 1) attaching an EEG cap to each subject's head, 2) displaying the text (reading materials) on a screen, and 3) recording the EEG and eye-tracking (for verification and calibration of the EEG signals) data while the subject is reading the text. The EEG signals are further extracted from the recorded data and fed as input to a decoding model to predict the original text the subject was reading on the screen.

Formally, this task can be formulated as a sequence-to-sequence machine translation task as follows:
\begin{equation}
P(Y | X) = \arg\max_Y \prod_{t=1}^{T'} P(y_t | y_{<t}, X)
\label{con:EEG to Text Translation}
\end{equation}
where $T'$ represents the length of the target sentence $Y$; $y_t$ represents the word or token at position $t$ in the target sentence $Y$; $y_{<t}$ represents the words or tokens preceding position $t$ in the target sentence $Y$; $X$ represents the input EEG data; and $P(y_t | y_{<t}, X)$ is the conditional probability of generating word $y_t$ given the previous words $y_{<t}$ and the input EEG data $X$. Our goal is to maximize the probability $P(Y | X)$ of generating the target sentence given the input EEG data.

\section{Methodology}
\begin{figure*}[t]
\centering
\includegraphics[width=\linewidth]{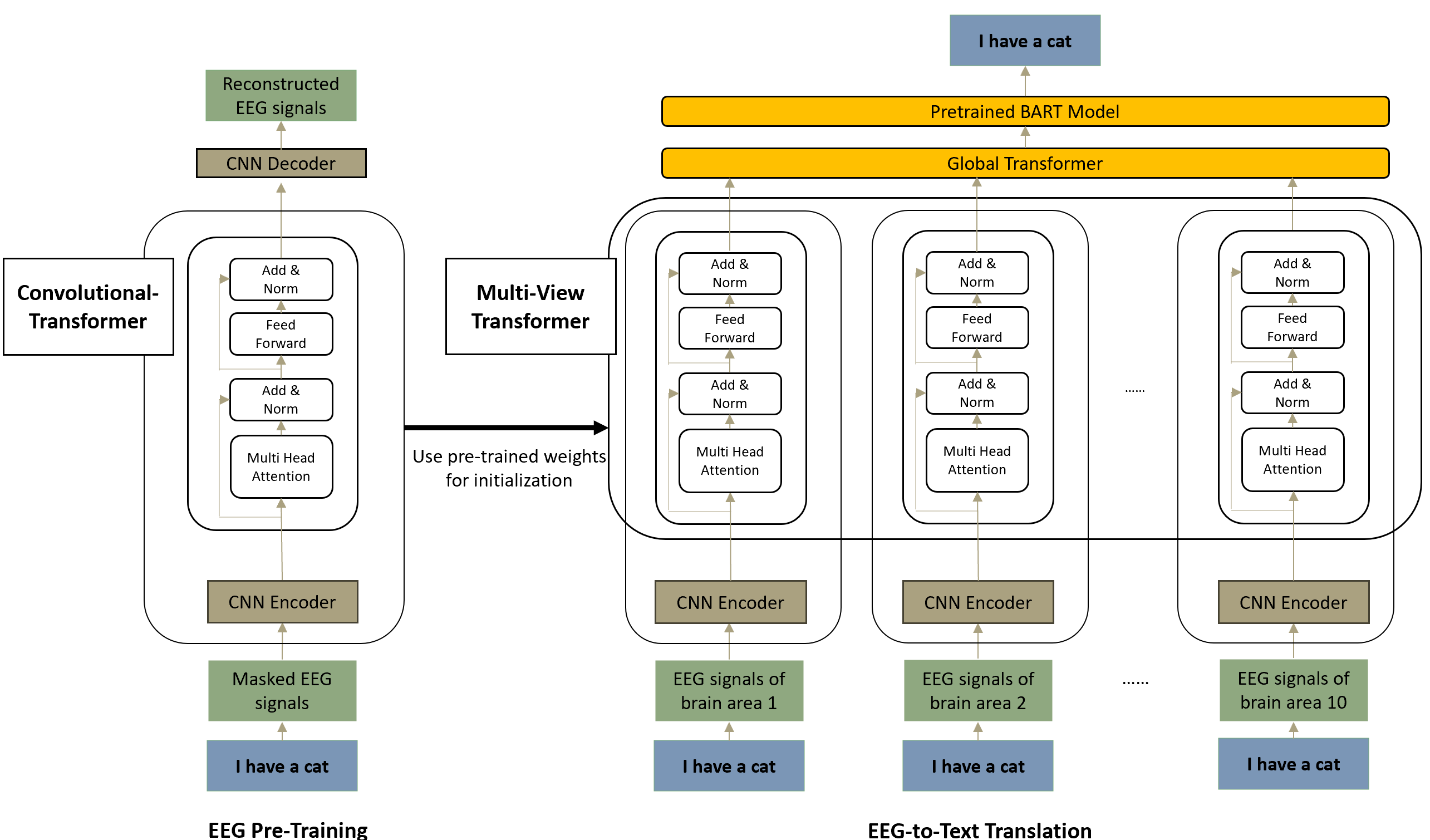}
\vspace{-4mm}
\caption {The overall framework of \our. It takes the sentence EEG signals as input and decodes the original text as output. \our includes major steps of 1) a base convolutional transformer model, 2) pre-training for EEG encoding, and 3) a multi-view transformer for different spatial regions of the brain.}  
\label{fig:main figure}
\end{figure*}

\subsection{Baseline Model}
Our baseline model \cite{DBLP:journals/corr/abs-2112-02690} takes the word-level EEG features as the input to a transformer model followed by a pre-trained BART model for text decoding. The raw EEG signals are typically stored as a two-dimensional array with one dimension for time and the other for channels (the number of electrodes used to collect EEG signals). Each value in this two-dimensional array corresponds to the signal strength collected at the corresponding time for the corresponding channel. 
In the baseline model, the word-level EEG features are extracted from eight independent frequency bands from the raw EEG signals. The above eight word-level EEG features are simply concated across all the channels as input to the decoder framework. 

The baseline model faces the following challenges: 1) the reliance on eye-tracking calibration for word-level EEG feature extraction introduces error propagation and lacks generalizability to scenarios such as inner speech decoding \cite{martin2018decoding,nalborczyk2020can}, 2) there is room for improvement in EEG representation learning through self-supervised pre-training, and 3) the lack of spatial resolution modeling ignores the varying importance of different brain regions in language processing. To overcome these challenges, we propose a novel framework, \our, that achieves superior performance for open-vocabulary EEG-to-text translation.

\subsection{Convolutional Transformer for Sentence-Level EEG Encoding}
Instead of using the word-level EEG features crafted based on the eye-tracking data, we directly use the sentence-level EEG signals as input to our model. Using sentence-level EEG signals offers several advantages over word-level EEG features. It provides richer information without error propagation from the eye-tracking data and exhibits better generalizability to other tasks, such as inner speech decoding, where acquiring eye-tracking data is infeasible.

However, the sentence-level EEG signals pose a challenge due to their excessive length, potentially overloading laboratory-level GPUs if directly input into the transformer layer. To tackle this issue, we introduce a convolutional transformer model that incorporates a CNN module for compressing raw EEG signals. This CNN module comprises two convolutional layers, adept at both temporal and spatial (or channel) compression. We also compared two input formats of the sentence-level EEG signals: 1) the raw signals, and 2) the spectrogram of the signals. The spectrogram of a signal (Appendix Figure \ref{fig:spec}) is a two-dimensional image, where the x-axis represents time, the y-axis represents frequency, and the image pixel value represents the magnitude of the signal at each time-frequency pair. The sentence-level EEG signals are then input into the CNN module to obtain compressed EEG signals, which are then fed into the transformer model for subsequent feature extraction and text translation.

\subsection{Transformer Pre-Training for an Enhanced EEG Encoding}

To enhance the semantic understanding of the EEG signals, we propose a self-supervised pre-training of the convolutional transformer model for parameter initialization (Figure \ref{fig:main figure}). Inspired by the masked language model pre-training strategies \cite{bert,spanbert,roberta}, we formulate our self-supervised pre-training objective as follows:
\begin{equation}
\theta^* = \arg\max_\theta \sum_{(i, j) \in \mathcal{D}} \log P(M | C; \theta),
\end{equation}
where $M$ represents the masked tokens; $C$ represents the context or surrounding tokens; $\theta^*$ represents the optimal model parameters; $\theta$ represents the model parameters being optimized; $\mathcal{D}$ represents the training data, where $(i, j)$ are pairs of sentences or sentence fragments; and $P(M | C; \theta)$ is the probability of predicting the masked tokens.

During the self-supervised pre-training stage, we add a convolutional decoder module on top of the convolutional transformer encoder to decode the input EEG signals. The input is the sentence-level EEG signals masked with different strategies and the output is the sentence-level EEG signals reconstructed by the CNN decoder. Specifically, we compared three different masking strategies for the sentence-level EEG signals as follows:
\begin{itemize}[leftmargin=*]
\item \textbf{Masked Token Prediction} \cite{bert}: randomly masking 15\% of all the tokens. 
\item \textbf{Continuous Masked Token Prediction} \cite{spanbert}: randomly masking a sequence of consecutive tokens until a total of 15\% of all the tokens are masked.
\item \textbf{Re-Masked Token Prediction }\cite{roberta}: re-randomizing the masking of 15\% of all the tokens for each training epoch.
\end{itemize}

It is important to highlight that our self-supervised pre-training step allows for seamless integration of EEG data from diverse tasks, including image recognition. In our experiments, we further incorporated an image EEG dataset \cite{imageeeg} during pre-training, aiming to showcase the model's adaptability to EEG signals from multi-modal data and explore the potential for enhanced translation performance through the combination of EEG signals from diverse data modalities.

The goal of this pre-training step is to have the convolutional transformer learn meaningful concepts such as context, relationships, and semantics present in sentence-level EEG signals during this pre-training process. After pre-training, the parameters are saved and used as the initial parameters for the final multi-view transformer model.

\begin{table}[t]
    \centering
    \small
    \begin{tabular}{m{11em}|m{14em}}
    \toprule 
    \multirow{1}{*}{Approximate Brain Areas} 
     & Corresponding Electrodes\\
     \midrule
     \multirow{1}{*}{Prefrontal Cortex} 
     & E6, E12, E5, E11, E16, E15, E20, E118, E24, E124, E26, E2, E27, E123 E3, E4, E23, E19, E22, E9, E10, E18, E28, E33, E117, E122\\
     \midrule
     
    \multirow{1}{*}{Premotor Cortex} 
     & CZ, E7, E106, E105, E104, E115, E114, E120, E110, E116, E121, E111, E112, E109, E13, E30\\
     \midrule

    \multirow{1}{*}{Broca's Area} 
     & E29, E36, E35, E34\\
     \midrule
     
    \multirow{1}{*}{Auditory Association Area} 
     & E40, E38, E39, E43, E44, E46, E57, E58, E64\\
     \midrule
    \multirow{1}{*}{Primary Motor Cortex} 
     & E31, E80, E55, E37, E87, E93, E103, E102, E108\\
     \midrule
    \multirow{1}{*}{Primary Sensory Cortex} 
     & E54, E79, E61, E78, E62, E53, E86, E92, E98, E100, E101\\
     \midrule
    \multirow{1}{*}{Somatic Sensory Cortex} 
     & E67, E77, E71, E72, E76, E66, E84, E60, E85\\
     \midrule
    \multirow{1}{*}{Auditory Cortex} 
     & E59, E91, E97, E51\\
     \midrule 
     
    \multirow{1}{*}{Wernicke's Area} 
     & E41, E42, E52, E47, E45, E50\\
     \midrule     
    \multirow{1}{*}{Visual Area} 
     & E65, E69, E70, E74, E75, E82, E83, E89, E90, E95, E96\\
     \midrule     
    \end{tabular}
            \caption{Ten channel groups and their corresponding approximate brain areas.}
    \label{tab:electreodes mapping}

\end{table}

\subsection{Multi-View Transformer for Different Spatial Regions of the Brain}
Another important feature of our model is the novel multi-view transformer decoder architecture we introduced that encodes different regions of the brain with a different convolutional transformer  (Figure \ref{fig:main figure}). The multi-view transformer model takes into account the fact that different brain regions potentially play different roles in language processing. This spatial modeling therefore can improve the model performance, but has been overlooked in previous work.

We partition the 105 channels into ten groups based on their spatial location under the guidance of functional brain regions (Table \ref{tab:electreodes mapping}). Specifically, we compared the spatial distribution of 105 electrodes with the spatial distribution of functional brain regions and mapped each electrode to its closest brain region. Details of the electrode spatial distribution can be found in \cite{hollenstein2018zuco}.

After the partition of the electrodes, we create a multi-view transformer model including ten convolutional transformers at the bottom level, where each convolutional transformer encodes the EEG signals from the electrodes in that region. On top of the ten convolutional transformers, we add a global transformer to unify the information from different brain regions. The combined information from the global transformer is further fed into the BART model for text decoding. 

In summary, the multi-view transformer envisions multiple parallel convolutional transformer models where each captures different aspects of EEG signals combined from different spatial regions of the brain regions. This approach enhances the spatial resolution of the model and further improves the text decoding performance.

\section{Experiment}
\subsection{Experimental Setup}
\paragraph{Dataset}

We utilize both the Zuco \cite{hollenstein2018zuco} and Image-EEG \cite{imageeeg} datasets for pre-training and use Zuco to train the multi-view transformer and BART model for text decoding. Details of both datasets are listed below.
\begin{itemize}[leftmargin=*]
    \itemsep0em
    \item \textbf{Zuco} \cite{hollenstein2018zuco} contains EEG and eye-tracking data from 12 healthy adult native English speakers engaged in natural English text reading for 4 - 6 hours. This dataset covers two standard reading tasks and a task-specific reading task, offering EEG and eye-tracking data for 21,629 words across 1,107 sentences and 154,173 fixations. We pre-process the data to extract both word-level and sentence-level features to serve as input to our model.
    \item \textbf{Image-EEG} \cite{imageeeg} is a large and rich dataset containing high temporal resolution EEG signals of images of objects on natural backgrounds. The dataset included 10 participants, each performing 82,160 trials across 16,740 image conditions.
\end{itemize}

\paragraph{Baselines}
We compare \our with two baseline models for open-vocabulary EEG-to-text translation.
\begin{itemize}[leftmargin=*]
    \itemsep0em 
    \item \textbf{Baseline} \cite{DBLP:journals/corr/abs-2112-02690} uses word-level EEG signals as input to a transformer model followed by a pre-trained BART model for decoding.
    \item \textbf{DeWave} \cite{duan2023dewave} introduces a discrete codex encoding after the transformer layer, and uses both word-level EEG features and the raw EEG signals as input. 
\end{itemize}

\paragraph{Evaluation Metrics} 
We utilize BLEU-1, BLEU-2, BLEU-3, BLEU-4, and ROUGE-1 evaluation metrics to compare the performance of \our with the baselines.

The BLEU-N scores (N = 1, 2, 3, 4) are used to measure the quality of the generated text, with higher values indicating better performance.
\begin{equation}
\text{BLEU} = \text{BP} \cdot \exp\left(\sum_{n=1}^{N} w_n \cdot \log\left(\frac{\text{count}_{\text{clip}, n}}{\text{count}_{\text{ref}, n}}\right)\right),
\end{equation}
where \text{BLEU} represents the BLEU score; \text{BP} represents the brevity penalty; $N$ represents the max n-gram order; $w_n$ represents the n-gram weights; $count_{clip, n}$ represents count of candidate n-grams in reference and $count_{ref, n}$ represents count of reference  n-grams.

ROUGE-1 scores, which include F (F1-score), P (precision), and R (recall), are used to evaluate the overlap between generated text and reference text. 

\begin{equation}
\text{ROUGE-1} = \frac{\sum_{\text{ref}} \sum_{\text{1-gram}} \min(\text{match}, \text{ref})}{\sum_{\text{ref}} \sum_{\text{1-gram}} \text{ref}},
\end{equation}
where \text{ROUGE-1} represents the ROUGE-1 score; \text{match} represents the count of matching 1-gram; \text{ref} represents the count of 1-gram.

\begin{table}[t]
    \centering
    \scriptsize
    \begin{tabular}{c c c}
        \toprule
         \textbf{Methods} &  \textbf{Batch Size} & \textbf{Learning Rate}  \\ \midrule

         \our (Convolutional Transformer) & 4 & $1\times 10^{-5}$ \\ \midrule
        \our (+ Pre-training) & 4 & $5\times 10^{-5}$ \\ \midrule
         \our (+ Multi-View Transformer) & 4 & $3\times 10^{-5}$ \\ \bottomrule
    \end{tabular}
        \caption{Optimal hyper-parameters for \our ablations.}
    \label{tab:hyper_params}
\end{table}

\begin{table*}[t]
  \centering
    \begin{tabular}{c cccc ccc}
    \toprule
     \multirow{2}{*}{\textbf{Methods}} &
     \multicolumn{4}{c}{\textbf{BLEU-N}} & \multicolumn{3}{c}{\textbf{ROUGE-1}} \\
     & N = 1 & N = 2 & N = 3 & N = 4 & F & P & R\\ 
    \midrule  
      Baseline \cite{DBLP:journals/corr/abs-2112-02690}  &  0.401 & 0.231  & 0.125 & 0.068 &  0.301 & 0.317 & 0.288  \\
      DeWave \cite{duan2023dewave} &  0.413  & 0.241 & 0.139 & 0.082 & 0.288 & 0.337 & 0.306 \\
    \midrule
      \our (Convolutional Transformer) & 0.400 & 0.236 & 0.137 & 0.082 & 0.325 & 0.361 & 0.297 \\
    \our (+ Pre-training)  & 0.445	& 0.274	&0.175&	0.117&	0.341&	0.383&	0.310\\
    \our (+ Multi-View Transformer) & \textbf{0.452} & \textbf{0.291} & \textbf{0.197} & \textbf{0.141} & \textbf{0.342} & \textbf{0.369} & \textbf{0.320} \\ 
        \bottomrule
    \end{tabular}%
  \caption{Performance comparison of \our with baseline methods.}
    \label{tab:our result}%
\end{table*}

\begin{table*}[t]
  \centering
    \begin{tabular}{c cccc ccc}
    \toprule
     \multirow{2}{*}{\textbf{Methods}} &
     \multicolumn{4}{c}{\textbf{BLEU-N}} & \multicolumn{3}{c}{\textbf{ROUGE-1}} \\
     & N = 1 & N = 2 & N = 3 & N = 4 & F & P & R\\ 
    \midrule  
          Spectrogram + Transformer   &  0.386 & 0.220  &  0.121 & 0.067  &  0.306 & 0.342 &  0.306\\
    Spectrogram + Convolutional Transformer & 0.374 & 0.209 & 0.112 & 0.061 &  0.302 & 0.339 &  0.274 \\

      EEG signal + Convolutional Transformer &  \textbf{0.400} & \textbf{0.236}  &  \textbf{0.137} & \textbf{0.082} &  \textbf{0.325} & \textbf{0.361} & \textbf{0.297} \\
        \bottomrule
    \end{tabular}%
      \caption{Ablation study of different input formats of the EEG signals.}    
  \label{tab:cnn ablation}%
\end{table*}

\paragraph{Parameter Study}
We used four A40 GPUs as our computing infrastructure and each training epoch took about 40 minutes.
The optimal hyper-parameters for our results are listed in Table \ref{tab:hyper_params}. The value ranges of each hyper-parameter are listed below:
\begin{itemize}[leftmargin=*]
\itemsep0em 
    \item Batch Size $\in \{4,8,16\}$

      \item Learning Rate $\in \{$1$\times 10^{-6}$, $3\times 10^{-6}$, $5\times 10^{-6}$, $7.5\times 10^{-6}$, $8\times 10^{-6}$, $9\times 10^{-6}$, $1\times 10^{-5}$, $2\times 10^{-5}$, $3\times 10^{-5}$, $4\times 10^{-5}$, $5\times 10^{-5}$, $7.5\times 10^{-5}$, $1\times 10^{-4}$, $3\times 10^{-4}$, $5\times 10^{-4}$, $7.5\times 10^{-4}$, $1\times 10^{-3}$\}
\end{itemize}

\begin{table*}[t] 
  \centering
    \begin{tabular}{c cccc ccc}
    \toprule
     \multirow{2}{*}{\textbf{Methods}} &
     \multicolumn{4}{c}{\textbf{BLEU-N}} & \multicolumn{3}{c}{\textbf{ROUGE-1}} \\
     & N = 1 & N = 2 & N = 3 & N = 4 & F & P & R\\ 
    \midrule  
    
      Masked Token Prediction  &  0.409 & 0.242  &  0.141 & 0.087  & 0.325 & 0.357 & 0.300\\
      Continuous Masked Token Prediction    &  0.411 & 0.243  & 0.137  & 0.078  &  0.319 & 0.352 & 0.294\\
     Re-Masked Token Prediction & \textbf{0.431} & \textbf{0.260} & \textbf{0.157} & \textbf{ 0.098} & \textbf{0.330} & \textbf{0.361} & \textbf{0.306} \\
        \bottomrule
    \end{tabular}%
      \caption{Ablation study of different pre-training strategies of the EEG signals.}  
  \label{tab:MLM ablation}%
\end{table*}

\begin{table*}[t]
  \centering
    \begin{tabular}{c cccc ccc}
    \toprule
     \multirow{2}{*}{\textbf{Methods}} &
     \multicolumn{4}{c}{\textbf{BLEU-N}} & \multicolumn{3}{c}{\textbf{ROUGE-1}} \\
     & N = 1 & N = 2 & N = 3 & N = 4 & F & P & R\\ 
    \midrule  
        Single-View without image-EEG   & 0.431 & 0.260 &  0.157 & 0.098 &  0.330 & 0.361 & 0.306 \\
        Single-View with image-EEG &  0.445	& 0.274	&0.175&	0.117&	0.341&	0.383&	0.310\\
        \midrule
       Multi-View  without image-EEG &  0.442 & 0.277 & 0.179 & 0.121 & 0.335 & 0.365 & 0.311\\ 
      Multi-View with image-EEG & \textbf{0.452} & \textbf{0.291} & \textbf{0.197} & \textbf{0.141} & \textbf{0.342} & \textbf{0.369} & \textbf{0.320} \\
        \bottomrule
    \end{tabular}%
      \caption{Ablation study of adding image-EEG data into pre-training.}  
  \label{tab:image eeg result}%
\end{table*}

\begin{table*}[t]
  \centering
    \begin{tabular}{c cccc ccc}
    \toprule
     \multirow{2}{*}{\textbf{Methods}} &
     \multicolumn{4}{c}{\textbf{BLEU-N}} & \multicolumn{3}{c}{\textbf{ROUGE-1}} \\
     & N = 1 & N = 2 & N = 3 & N = 4 & F & P & R\\ 
    \midrule  
      Only Global Transformer   &  0.404	&0.238&	0.139&	0.084&	0.303	&0.335	&0.279\\
      + One Convolutional Transformer &  0.436	&0.270&	0.168	&0.110&	 0.327	&0.363	&0.299\\
      + Three Convolutional Transformers & \textbf{0.442} & \textbf{0.277} & \textbf{0.179} & \textbf{0.121} & \textbf{0.335} & \textbf{0.365} & \textbf{0.311} \\ 
        \bottomrule
    \end{tabular}%
      \caption{Ablation study of different training strategies of the multi-view transformer.} 
  \label{tab:multi view result 1}%
\end{table*}

\begin{table*}[t]
    \centering
    \begin{tabular}{m{1em}|m{35em}}
    \toprule 
     
     \multirow{3}{*}{(1)} 
     & Ground Truth:  It's not a particularly \textbf{good film}, but neither is it a \textbf{monsterous} one.\\
     \cmidrule{2-2}
     & Baseline Output: was a a bad \textbf{good} story, but it is it \textbf{bad bad}. one.\\
    \cmidrule{2-2}
     & \our output: 's a a \textbf{great} romantic \textbf{movie}, but it is it the \textbf{disaster} movie one.\\
     \midrule    
     \multirow{3}{*}{(2)} 
     & Ground Truth: He won a \textbf{Nobel Prize in Chemistry} in 1928\\
     \cmidrule{2-2}
     & Baseline Output: was the Pulitzer Prize for Literature in 18.  \\
    \cmidrule{2-2}
     & \our Output: won \textbf{Nobel Prize in Chemistry} for 1901 \\
     \midrule
          \multirow{3}{*}{(3)} 
          & Ground Truth: The book was awarded the 1957 \textbf{Pulitzer Prize for Biography}.\\
     \cmidrule{2-2}
     & Baseline Output: first is published the Pulitzer \textbf{Pulitzer Prize} for Fictionography.  \\
    \cmidrule{2-2}
     & \our Output: book is a \textbf{Pulitzer Prize for Biography}. \\
    \bottomrule
    \end{tabular}
        \caption{Case study of the output sentences comparing \our and the baseline method \protect\cite{DBLP:journals/corr/abs-2112-02690}.}
    \label{tab:generation_results}
\end{table*}

\subsection{Results}

\paragraph{Main Results}
Table \ref{tab:our result} shows our main experimental results. 
The baseline method \cite{DBLP:journals/corr/abs-2112-02690} achieves a moderate performance in text decoding with BLEU scores. DeWave \cite{duan2023dewave} slightly improved the performance across all metrics, demonstrating the effectiveness of discrete encoding. \our improved the text decoding performance by a large margin due to several technical innovations. First, a single convolutional transformer achieved slightly lower BLEU scores (BLEU-1: -1.3\%; BLEU-2: -0.5\%; BLEU-3: -0.2\%; BLEU-4: -0.0\%) but higher ROUGE-1 scores (F1-score: +3.7\%; Precision: +2.4\%; Recall: -0.9\%) compared to DeWave.
Second, \our with pre-training further enhanced the BLEU scores (BLEU-1: +1.8\%; BLEU-2: +1.9\%; BLEU-3: +1.8\%; BLEU-4: +1.6\%)  and ROUGE-1 scores (F1-score: +4.2\%; Precision: +2.4\%; Recall: +0.0\%) compared to DeWave. Pre-training proved effective in enhancing text generation by providing a strong initialization foundation for our model.
Third, \our with multi-view transformers achieved the highest scores across all metrics, with a significant increase in the BLEU scores (BLEU-1: \textbf{+3.9\%}; BLEU-2: \textbf{+5.0\%}; BLEU-3: \textbf{+5.8\%}; BLEU-4: \textbf{+5.9\%}) and ROUGE-1 scores (F1-score: \textbf{+5.4\%}; Precision: \textbf{+3.2\%}; Recall: \textbf{+1.4\%}) compared to DeWave. \our excelled in generating coherent, contextually relevant, and high-quality text.  

\paragraph{Convolutional Transformer}
We first compare different input representations of the EEG signals to see how the representation affects the performance of a base convolutional transformer model. In this ablation study, we compare the raw EEG signals with their spectrograms using the fast Fourier transform \cite{cochran1967fast} to convert the original one-dimensional time array into a two-dimensional time-frequency matrix. The results are shown in Table \ref{tab:cnn ablation}. Using the raw EEG as the input consistently led to better performance than using the spectrogram as the input. Because the spectrogram only keeps the magnitude information and ignores the phase information of the raw EEG signal, the superior performance of the raw EEG signal suggested that the phase information might be important for decoding. Therefore, the raw EEG signals are used as the input in our subsequent experiments.  

\paragraph{EEG Pre-Training}
We then conducted ablation experiments to compare the effectiveness of three pre-training strategies: 1) Masked Token Prediction \cite{bert}, 2) Continuous Masked Token Prediction, and 3) Re-Masked Token Prediction \cite{roberta}. The results are shown in Table \ref{tab:MLM ablation}.
The Re-Masked Token Prediction \cite{roberta} exhibits the best performance among all the three masking strategies. 
One potential reason is that the convolutional transformer model can learn more diverse semantic information by masking different tokens in each training epoch during pre-training.

In the above study, we focused on identifying the optimal pre-training strategy among the three without incorporating image-EEG data \cite{imageeeg}. As an additional component, we introduced image-EEG data to assess the compatibility of our model with EEG signals from multi-modal inputs. Leveraging our self-supervised pre-training strategy, we directly incorporated image-EEG data into the pre-training phase to enable the model to glean knowledge from diverse sources. The results, detailed in Table \ref{tab:image eeg result}, demonstrate that adding image-EEG data significantly enhances translation performance for both the single convolutional transformer and the multi-view transformer.

\paragraph{Multi-View Transformer}
Finally, we compare different training strategies of the multi-view transformer to demonstrate the effectiveness of the multi-view transformer and find the best training strategy. The image-EEG data was not included in this ablation study. Specifically, we compared three training strategies as follows:
\begin{itemize}[leftmargin=*]
\item \textbf{Only Global Transformer}: Fixing the parameters of all 10 convolutional transformer modules and training only the global transformer for text decoding.
\item \textbf{Global Transformer + One Convolutional Transformer}: During each training epoch, train one convolutional transformer with the global transformer while fixing the parameters of the remaining nine convolutional transformers.
\item \textbf{Global Transformer + Three Convolutional Transformers}: During each training epoch, train three convolutional transformers with the global transformer while fixing the parameters of the remaining seven convolutional transformers.
\end{itemize}

The results in Table \ref{tab:multi view result 1} demonstrate that activating three convolutional transformers together with the global transformer achieves the best performance.

This suggests further improvement may be attainable by increasing the number of activated convolutional transformers during each training epoch if more GPU resources are available.

\paragraph{Case Study}
Table \ref{tab:generation_results} shows our case study results. 
In the first sentence, the baseline model accurately translates "good," whereas \our, in addition, accurately captures the first half of the sentence with "movie" (synonymous with "film"). Additionally, \our correctly translates the second half of the sentence with "disaster movie" corresponding to "monstrous one" in the original sentence. 
In the second sentence, \our accurately captured "won Nobel Prize in Chemistry," while the baseline produced incorrect information, stating "Pulitzer Prize" and the wrong field, "Literature."
In the third sentence, both \our and the baseline correctly identified "book" and "Pulitzer Prize." However, \our, in addition, correctly identified the field as "Biography," while the baseline erroneously outputted "Fictionography."

In addition, we conducted an interesting case study to show that \our has the ability of zero-shot image-to-text translation. Details can be found in Appendix \ref{sec:image-to-text}.

\section{Related Work}
\label{sec:bibtex}
\paragraph{Brain Computer Interface}
The landscape of brain-to-speech and brain-to-text decoding encompasses three principal approaches grounded in the features they capture: motor imagery-based, overt speech-based, and inner speech-based. These methods explore a variety of brain signals, including electroencephalogram (EEG), electrocorticography (ECoG), and functional magnetic resonance imaging (fMRI). Despite these endeavors, existing approaches exhibit limitations concerning vocabulary size, articulation dependence, speed, and device compatibility. Motor imagery-based systems, exemplified by point-and-click \cite{pandarinath2017high} mechanisms and imaginary handwriting \cite{willett2021high}, showcase high accuracy but modest typing rates. Overt speech-based techniques for decoding or synthesizing speech offer expedited communication rates. However, they require either physical vocal tract movement \cite{herff2015brain,anumanchipalli2019speech,makin2020machine} or mental articulation imagination \cite{moses2021neuroprosthesis,willett2023high}. This engenders language dependency and pronunciation variations across languages. Another line of research tackles articulation dependency by decoding imagined speech \cite{nieto2022thinking} or reading text \cite{sun2019towards,panachakel2021decoding}. 

Our work follows this line of decoding reading text directly from EEG signals. 

\paragraph{EEG-to-Text Decoding}
Prior investigations into the decoding of EEG-to-text, as documented in the literature \cite{herff2015brain,sun2019towards,anumanchipalli2019speech,makin2020machine,panachakel2021decoding,moses2021neuroprosthesis,nieto2022thinking}, have demonstrated commendable accuracy when applied to limited and closed vocabularies. Nevertheless, these studies encounter challenges in attaining comparable levels of accuracy when confronted with more extensive and open vocabularies. New investigations have expanded their focus from closed-vocabulary EEG-to-text decoding to encompass open-vocabulary scenarios \cite{DBLP:journals/corr/abs-2112-02690,willett2023high,tang2023semantic,duan2023dewave}. The two research studies most similar to our work are a baseline method \cite{DBLP:journals/corr/abs-2112-02690} and DeWave \cite{duan2023dewave}. The baseline method proposes a framework utilizing transformer and pre-trained BART language models, which establish baseline performance of open-vocabulary EEG-to-text translation. DeWave employs a quantization encoder to derive discrete encoding and aligns it with a pre-trained language model for the open-vocabulary EEG-to-text translation.
The limitations of both the baseline method and DeWave lie in their reliance on eye-tracking calibration for word-level EEG feature extraction that introduces error propagation and lacks generalizability to scenarios such as inner speech decoding. \our improves the open-vocabulary EEG-to-text translation performance as well as enhancing the generality by requiring only sentence-level EEG signals as input.

\paragraph{EEG Encoding}
It is a challenging problem to effectively encode the long and noisy EEG signals to facilitate subsequent decoding tasks. In Conformer \cite{song2022eeg}, the authors propose a compact convolutional transformer, named EEG Conformer, to encapsulate local and global features in a unified EEG classification framework. Specifically, the convolution module learns the low-level local features throughout the one-dimensional temporal and spatial convolution layers. The self-attention module is straightforwardly connected to extract the global correlation within the local temporal features. However, in the case of the Conformer model, the authors trained this model from scratch, whereas \our further incorporated pre-training and multi-view settings to enhance the text translation performance.

\paragraph{EEG Pre-Training}
Recent work, such as BrainBERT \cite{wang2023brainbert}, BENDR \cite{kostas2021bendr} and MAEEG \cite{maeeg}, has been done on EEG signal pre-training that greatly inspired \our. 

BrainBERT converts intracranial recordings to spectrograms and uses spectrograms as input. BrainBERT then masks multiple continuous bands of random frequencies and time intervals from spectrograms and aims to reconstruct the original spectrogram. BENDR uses raw EEG signals as input. After a convolutional layer, the raw EEG signals are converted to embedding features. These embedding features are masked by using masked token prediction \cite{bert} and the reconstruction goal is the original embedding features.
MAEEG uses raw EEG signals as input and masks the embedding features of the convolutional layer generated with a masked token prediction as BENDR. However, MAEEG's reconstruction goal is the raw EEG signals. \our directly masks the raw EEG signals with the pre-training objective to reconstruct the raw EEG signals. \our also experimented with various masking strategies and incorporated EEG signals for the pre-training process.

\section{Conclusion}

In this work, we proposed a novel EEG-to-text decoding model, \our that takes raw EEG signals as input and leverages EEG pre-training and a multi-view transformer to enhance the decoding performance. \our achieved superior performance for open-vocabulary EEG-to-text decoding. Future work includes expanding the model's capabilities to EEG signals from diverse multi-modal data.

\clearpage

\section*{Ethics Statement}
Given our current methodology design, we believe that no significant ethical concerns are likely to arise. We have diligently used openly accessible datasets and models, enhancing transparency and accessibility in our study on EEG signal processing - a task gaining attention in Brain-Computer Interface (BCI) research.

However, it is crucial to note that our architectural framework relies on the pre-trained model, BART, which may make biased decisions influenced by its pre-training data. While our experiments have not shown explicit performance issues due to biases, we must recognize that this observation may be limited to the specific dataset and pre-trained model we used. It is essential to stay vigilant and continue exploring methods to mitigate and correct potential biases that could arise when using pre-trained models. 

\bibliographystyle{named}
\bibliography{eeg}

\clearpage

\appendix

\section{EEG to Stectrogram}
Figure \ref{fig:spec} shows a piece of EEG signals and its corresponding spectrogram.
\setcounter{figure}{0}
\renewcommand{\thefigure}{A\arabic{figure}}
\begin{figure*}[t!]
\centering
\includegraphics[width=\linewidth]{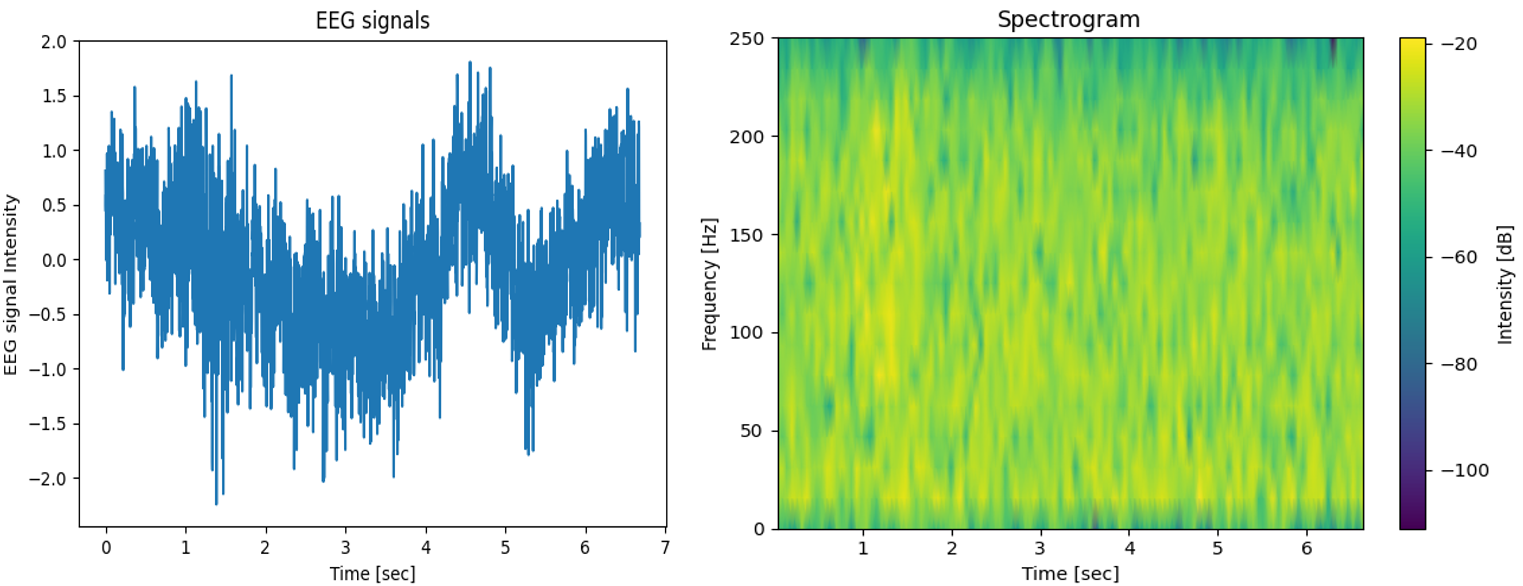}
\vspace{-4mm}
\caption {a piece of EEG signals and its corresponding Spectrogram}  
\label{fig:spec}
\end{figure*}

\section{Zero-Shot Image-to-Text Translation}
\label{sec:image-to-text}
\begin{figure*}[t]
     \centering
     \begin{subfigure}[b]{0.4\textwidth}
         \centering
         \includegraphics[width=\textwidth]{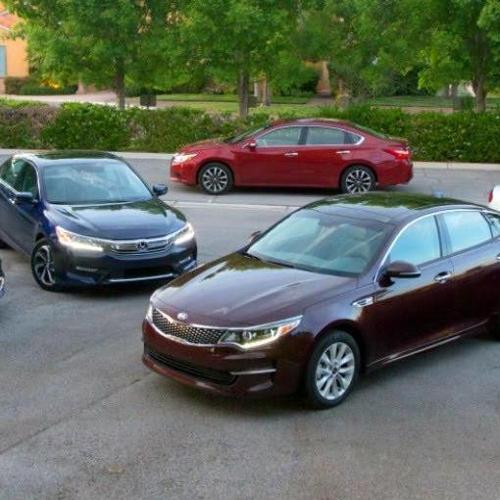}
         \caption{An image of car. The translation result of \our is: "alog,,. \textbf{car},,,,,,,,,,,,,,,,,,,,,,"}
         \label{car}
     \end{subfigure}
     \hfill
     \begin{subfigure}[b]{0.4\textwidth}
         \centering
         \includegraphics[width=\textwidth]{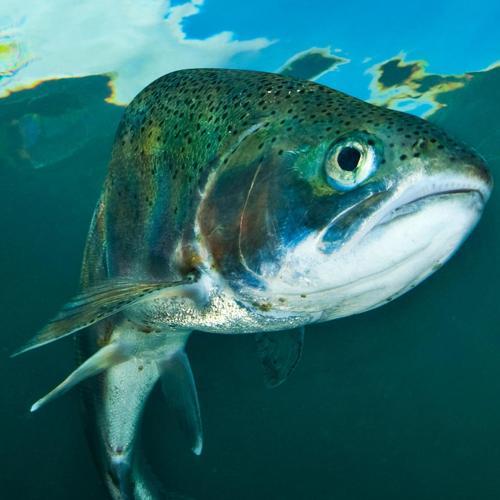}
         \caption{An image of car. The translation result of \our is: "\textbf{fish},,... has,,,,,,,,,,,,,,,,,,,,,,,,,,"}
         \label{fish}
     \end{subfigure}
     \caption{Zero-Shot Image-to-Text Translation.}
    \label{image-to-text}
\end{figure*}

Figure \ref{car} and \ref{fish} show the zero-shot image-to-text translation results. We directly input the EEG signals of image-EEG data into the multi-view transformer model after training, and the output results are image-to-text translation results. The first image contains multiple cars, and the output accurately captures the "car" keyword. The second image contains a fish, and the output captures the "fish" keyword equally accurately.

\end{document}